\documentclass[arxiv]{latex/melba}

\usepackage{mwe} 

\usepackage{amsmath,amsfonts}

\usepackage{multirow}
\usepackage[table,xcdraw]{xcolor}
\usepackage{adjustbox}
\usepackage{subcaption}
\usepackage{dblfloatfix}

\melbaid{2026:022}  
\doi{https://doi.org/10.59275/j.melba.2026-3bfa}
\melbaauthors{I. Islam, B. Ruijsink, E. Puyol-Ant\'{o}n, A. J. Reader and A. P. King}
\email{iman.islam@kcl.ac.uk}
\volume{2026}
\firstpageno{464}  
\melbayear{2026}  
\datesubmitted{11/2024}  
\datepublished{07/2026} 

\ShortHeadings{Learning with Partially Labelled Data from Multiple Domains}{I. Islam}

\title{Comparison of Loss Functions for Robust Deep Learning-based Echocardiography Segmentation when Learning with Partially Labelled Data from Multiple Domains}

\author{\name Iman Islam\aff{1}\orcid{0009-0002-0856-5498}, \name Bram Ruijsink\aff{1}\orcid{0000-0001-8313-5709},
        \name Esther Puyol-Ant\'{o}n\aff{1}\orcid{0000-0002-9789-6629},
        \firstname Andrew J.\surname Reader \aff{1}\orcid{0000-0002-2726-3383},
	\name Andrew P. \surname King \aff{1}\orcid{0000-0002-9965-7015}
}
\affiliations{
	\num 1 \addr School of Biomedical Engineering {\&} Imaging Sciences, King's College London, UK \\
}

\abstract{
	Echocardiography is the first imaging modality used for assessing cardiac function, and accurate segmentation of cardiac structures is essential for deriving biomarkers. However, the development of effective automated segmentation models for multiple cardiac structures is challenged by the difficulty of training on datasets from different sources that are often partially-labelled. This study aims to address this challenge by evaluating the performance of three loss functions - adaptive categorical cross entropy (aCCE) loss, marginal loss, and the adaptive binary cross entropy (aBCE) loss - in handling partially-labelled data. We conduct a comprehensive comparison of these loss functions across multiple scenarios and network architectures: intra-domain and inter-domain tasks, with both single and multiple partial-labels, and varying proportions of fully-labelled to partially-labelled data.
    Our experiments reveal that all three loss functions exhibit strong performance in intra-domain segmentation tasks, effectively handling label variations within the same domain. For inter-domain tasks, where models are trained on datasets with a domain shift, the aBCE and marginal losses show superior performance when dealing with the case of one label being missing from some training examples. In scenarios involving more than one label being missing, marginal loss outperforms the other methods, demonstrating its robustness in such complex conditions. These results highlight the strengths of each loss function depending on the labelling scenario, emphasizing the importance of selecting the appropriate loss function to optimize model performance. This study represents the first investigation of techniques for handling partially-labelled data from multiple different domains in echocardiography segmentation and provides a comprehensive comparison of loss-based solutions.}


\begin{document}

\twocolumn[\maketitle]

\section{Introduction}
Echocardiography (echo) is typically the first imaging examination carried out when assessing cardiac function and can provide insight into the function of the heart chambers and valves. Useful biomarkers can be estimated by segmenting echo images and computing volumes of the cardiac structures. A common biomarker that is extracted is the left ventricular (LV) ejection fraction (EF), which can be calculated from the LV end-diastolic volume (EDV) and end-systolic volume (ESV). LV EF is routinely used for diagnosing and characterising heart failure \citep{bozkurt2021universal}. Other useful biomarkers include the left atrium (LA) volume, which can be a robust predictor of diastolic heart failure and is often analysed for patients with atrial fibrillation \citep{taniguchi2019usefulness}, and LV myocardial (LVM) mass, which can be used to detect cardiomyopathy \citep{berman2019physiology}. 


The analysis of echo examinations to estimate these biomarkers has traditionally been performed in a manual or semi-automatic manner.
However, echo images are challenging to segment due to the relatively low signal-to-noise ratio and the poor contrast between the blood pool and myocardium. Furthermore, echo images can suffer from artefacts such as intensity inhomogeneities as well as differences in image characteristics across patients and scanners.
Therefore, manual segmentation can be very time-consuming and is subject to intra- and inter-observer variability, which results in the biomarkers being poorly standardised across different annotators \citep{armstrong2015quality}.

Deep learning models have been proposed to automate the segmentation of echo images \citep{leclerc_deep_2019,madani_deep_2018,ghorbani_deep_2020,puyol2022ai} and a number of commercial solutions now exist, such as
GE Healthcare's
Vivid Cardiovascular Ultrasound
and 
Ultromics'
EchoGo.
However, for such tools to be more widely used in clinics across the world, which will have significant variations in terms of imaging equipment, operator expertise and patient populations, it is important that the models are robust to the variations in image characteristics that will be encountered. In other applications, such as cardiac magnetic resonance, segmentation models have been trained using diverse data sources and shown to be robust to such variations \citep{mariscal-harana_artificial_2023}. In this work we aim to address the methodological challenges involved in training a similarly robust model for echo segmentation.

One significant challenge that must be overcome when training models to segment multiple structures with diverse datasets is the variation in label presence in the training data. For instance, a number of public datasets exist for training echo segmentation models but the manually defined labels present are different. A summary of the most commonly used datasets can be found in Table \ref{tab:datasum} with a sample from each dataset shown in Figure \ref{fig:sample}. This variation in label presence means that the combination of the datasets will be \emph{partially-labelled}, i.e. not all structures will be labelled in all training samples. Training naively with partially-labelled datasets such as this can cause a conflict in the supervision due to the same structures being labelled as foreground in some samples and background in others. This paper focuses on the challenge of training a robust echo segmentation model using diverse partially-labelled data. We perform a comprehensive evaluation of different proposed solutions to this challenge and make recommendations for future development in this field.

\section{Related Works}
\label{sec:related}

A number of methods have been proposed to deal with partially-labelled data when training deep learning segmentation models, either by modifying the network architecture, its loss function or the training data to deal with the conflict in supervision. In addition, there have been differences in the definition of the problem being addressed. Below we review the proposed approaches, highlighting their commonalities and differences. 

The first class of methods aims to train a model using partially-labelled data to segment multiple \emph{non-overlapping} regions. When the same structure appears in multiple datasets it is considered to be a \emph{single} common structure, i.e. the segmentation model will only output a single label for this structure at inference time. \cite{petit_handling_2018} devised an early example of this class of method. They employed the Deeplab architecture and changed the loss function from cross entropy to a summation of binary cross entropy (BCE) losses, one for each segmentation class. For samples with missing labels, only the losses for the classes present in the ground-truth (GT) labels were computed. The method was demonstrated on a dataset of abdominal 3D computed tomography (CT) images.

\cite{roulet19a} investigated the same problem definition as \cite{petit_handling_2018}. They aimed to train a model on two brain magnetic resonance imaging (MRI) datasets, one of which included segmentations for the brain anatomy and one which had segmentations for brain lesions. In the brain lesion dataset, the rest of the brain was labelled as background. They proposed a method which  featured an adaptive cross entropy loss function to deal with the partial labelling which unifies the missing labels into a unique class. Their method employed a standard cross entropy loss when dealing with voxels within the brain lesion and the adaptive cross entropy loss when dealing with voxels outside the lesion. The loss function sums the network predictions for all non-lesion classes before the loss calculation.

\cite{shi_marginal_2021} also tackled the same problem definition and proposed a marginal loss function, which was similar in concept to the work of \cite{roulet19a}. They employed partially-labelled abdominal CT datasets and proposed a loss function in which any missing label in the training sample was merged with the background class before the loss calculation. In these merged regions, the loss function considered both the missing label and the background label to be a correct prediction. Therefore, the model was not penalized for predicting a label as foreground even if it was background in the ground truth segmentations.

Similarly, \cite{fidon2021label} also addressed the same problem in a similar way to \cite{roulet19a}. They proposed a loss function called the label-set loss and demonstrated it on fetal brain MRI images. The label-set loss function assumes that every voxel in the image has a set of potential labels that it could be. The loss is calculated based on the probability distribution over the possible labels in the label set.

More recently, the same problem was tackled by \cite{mariscal-harana_artificial_2023}, who trained a segmentation model on partially-labelled short-axis (SAX) cardiac magnetic resonance (CMR) images. The nnU-Net framework was employed and the cross entropy loss function was modified to not include model predictions of labels that were not present in the ground-truth. This approach was similar in concept to that of \cite{petit_handling_2018} but used categorical cross entropy (CCE) rather than BCE as the loss function. It was demonstrated on multiple large scale cine SAX CMR datasets in which one label (the LVM) was missing from a subset of the data (those representing the end systole frame).

\cite{dong_towards_2022} also tackled the problem of making single predictions of non-overlapping classes. However, this work aimed to exploit training data of a particular type. Their training data consisted of three chest X-Ray (CXR) datasets, each one containing a segmentation of a different organ, i.e. each dataset contained a single class. The aim was to train a model that could segment all three organs. The method worked by randomly sampling an image from each dataset and creating augmented images using MixUp, resulting in a new image with annotations of all three labels. These data were then used to train a fully-supervised segmentation model.

A final example of tackling this problem definition was reported by \cite{zhou_prior-aware_2019}, who proposed a method for the segmentation of abdominal CT images using partially-labelled datasets.  The training algorithm began with training the model on a small fully-labelled dataset which was then used to compute a prior distribution of the expected size of each organ. A compound loss of cross entropy and a prior loss was then employed when training with the full dataset, in which the prior loss ensured that the size distribution of predicted segmentations for each organ was consistent with that seen in the fully-labelled dataset. The model was then fine-tuned further on the partially-labelled dataset using this prior distribution information. For this method, a fully-labelled dataset is required to initialise the prior distributions.

An alternative problem definition was tackled by \cite{ulrich2023multitalent}. In this work, the authors aimed to learn differences in annotation styles between datasets as well as differences in label presence. Therefore, the same structure annotated in different datasets would be considered to be \emph{distinct} output labels of the model, each capturing the annotation style of the dataset. They proposed a solution called MultiTalent, in which the network architecture and the loss function was modified in order to predict all the desired labels. 
The loss function was a combination of sum of BCE losses (which was similar in concept to  \cite{petit_handling_2018}) and Dice loss for the labels that were present in the image. The method was demonstrated on segmentation using multiple partially-labelled CT datasets.

\cite{billot2024network} also tackled the same problem definition as \cite{ulrich2023multitalent} but modified the network architecture rather than the loss function. They proposed a conditioning network called CoNeMOS and used the U-Net as their base architecture. They placed label-conditioned layers (FiLM layers) after every convolutional layer. The parameters for these layers were chosen for each convolution using a multi-layer perceptron (MLP). 

\section{Contributions}

In this paper we tackle the first problem listed above, i.e. we consider a structure annotated in multiple datasets to be a single label, and we seek to produce non-overlapping segmentations. We extend our previous work \citep{mypaper} and compare the performance of state-of-the-art approaches to tackle this problem by analysing their viability when training models for echo segmentation using multiple diverse datasets with variable labelling. The loss functions we consider are the adaptive BCE loss by \cite{petit_handling_2018}, the marginal loss by \cite{shi_marginal_2021} and the adaptive CCE loss by \cite{mariscal-harana_artificial_2023}.

Our contributions are:
\begin{enumerate}
    \item We perform the most extensive quantitative comparison to date of different loss functions proposed for addressing the partial labelling problem in deep learning-based segmentation models.
    \item We perform the first evaluation of methods for dealing with partial labels in deep learning-based echo segmentation.
    \item We analyse the performance of the partial labelling loss functions across three different model architectures.
    \item We report the first systematic investigation of the impact on model performance of the relative numbers of fully-labelled and partially-labelled data in the training set.
    \item We investigate the performance of the different loss functions when dealing with single or multiple missing labels. 
\end{enumerate}

\section{Methods}
\subsection{Problem Definition}

Given $K$ datasets, $\mathcal{D}_k = \{(x,y)_s\}$, where $1 \leq k \leq K$, each dataset $\mathcal{D}_k$ is composed of $s$ image and GT segmentation pairs $(x,y)_s$ where $x$ is the image and $y$ is the GT. $y$ includes a multi-channel one-hot vector with $N$ channels for every pixel in the image. Each $\mathcal{D}_k$ can consist of a different number $s$ of image/GT pairs. $\Omega_N=\{1,...,N\}$ is the set of labels for the structures of interest ($N=3$ for our experiments).  For every $\mathcal{D}_k$ there is a corresponding set of labels $\mathcal{L}_k$ that are segmented in $y$. $\mathcal{L}_k$ is either equal to $\Omega_N$ (if the dataset is fully-labelled) or is a subset of it, i.e. $\mathcal{L}_k\subseteq \Omega_N$.  The channels in $y = \{y_{n}\}_{n=1}^{N}$ contain binary annotations if $n \in \mathcal{L}_k$, and zeroes otherwise. Our aim is to train a network $F_\theta$ (with trainable parameters $\theta$) to predict a single class for each pixel in each image $\hat{y} = F_\theta (x)$ (i.e. the predictions are non-overlapping).

\subsection{Datasets}
\label{sec:datasets}

\begin{figure}[t]
    \centering
    \includegraphics[width=\linewidth]{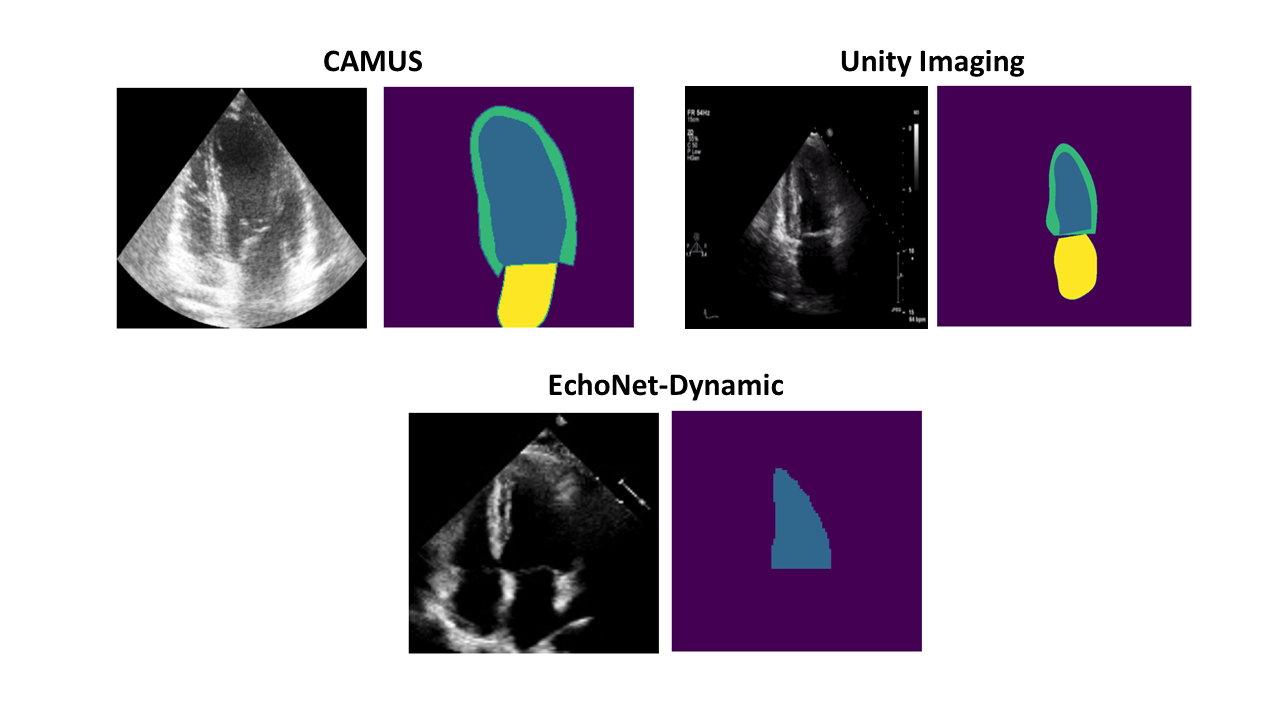}
    \caption{A sample image and ground truth segmentation from each of three public annotated echo datasets: CAMUS, Unity Imaging and EchoNet-Dynamic. Blue = LV, Green = LVM, Yellow = LA.}
    \label{fig:sample}
\end{figure}

Three publicly available datasets were exploited in this work: CAMUS \citep{leclerc_deep_2019}, EchoNet-Dynamic \citep{ouyang_video-based_2020} and Unity Imaging \citep{huang_fix--step_2023}. Example images from the three datasets can be seen in Figure \ref{fig:sample} and a summary of the number of samples and available GT labels is provided in Table \ref{tab:datasum}. 
The CAMUS and Unity Imaging datasets are fully-labelled and include segmentations for all three structures of interest (LV, LVM and LA) while the EchoNet-Dynamic dataset includes segmentations for the LV only. Further details of each dataset are provided below:

\label{sec:data}
\begin{itemize}
    \item \textbf{CAMUS} \citep{leclerc_deep_2019}: This dataset consists of apical 4- and 2-chamber images from 500 patients. These were acquired using GE Vivid E95 ultrasound scanners (GE Vingmed Ultrasound, Horten Norway) with a GE M5S probe (GE Healthcare, US). The segmentation masks for the LV, LVM and LA are available at end diastole (ED) and end systole (ES).
    \item \textbf{EchoNet Dynamic} \citep{ouyang_video-based_2020}: This dataset consists of apical 4-chamber and 2-chamber images from 10,024 individuals. The images were acquired using iE33, Sonos, Acuson SC2000, Epiq 5G, or Epiq 7C ultrasound machines. Segmentation masks for the LV at ED and ES and \revision{EF values for each patient are available.}
    \item \textbf{Unity Imaging} \citep{huang_fix--step_2023}: This dataset consists of apical 4- and 2-chamber images with ground truth delineations of the LV, LVM and LA for 1504 images. The delineations were provided as lists of coordinates and we converted these to segmentation masks. Upon manual review, several of these delineations were discarded due to the LVM being overlabelled into the right ventricular myocardium. Therefore, after excluding these we utilised 400 mostly apical 2-chamber images.
\end{itemize}
In addition, for external validation we employ one private clinical dataset, as described below:
\begin{itemize}
    \item \textbf{GSTFT Echocardiography Database}: This database contains A2C and A4C echocardiography data acquired from a cohort of heart failure with reduced ejection fraction (HFrEF) patients imaged at Guy’s and St. Thomas’ Foundation Trust (GSTFT).
    The ultrasound machines used were Philips IE33 and EPIQ 7C (Phillips Medical Systems, Andover, MA, USA) and the GE Vivid E9 (General Electric Medical Health, Milwaukee, WI, USA), each equipped with a matrix array transducer. This dataset has manual annotations of the LV, LVM and LA produced by a trained cardiologist.
\end{itemize}

\begin{table}[t]
    \centering
    \caption{Summary of the three datasets used in this work showing the number of subjects, number of images, ground truth segmentation labels present and image views. LV = left ventricle, LVM = left ventricular myocardium, LA = left atrium, A2C = apical 2-chamber, A4C = apical 4-chamber.}
    \resizebox{\linewidth}{!}{%
    \begin{tabular}{lcccccc} \cline{1-7}
         &  No. of subjects &  No. of images &  LV&  LVM&  LA& Image view(s) \\ \cline{1-7}
         CAMUS  &  500&  1000&  \checkmark&  \checkmark&  \checkmark& A2C + A4C \\  
         Unity Imaging &  - &  400&  \checkmark&  \checkmark&  \checkmark& A2C \\ 
         EchoNet-Dynamic   &  10024&  20048&  \checkmark&  &  & A2C + A4C \\ \hline
    \end{tabular}
    }
    \label{tab:datasum}
\end{table}

\subsection{Dataset Pre-processing}

Both the Unity Imaging dataset and the EchoNet-Dynamic dataset contained some markings or annotations outside of the cone. These markings were removed by using an nnU-Net model trained to segment the ultrasound cone. Following this, all images were resized to 256x256.

\subsection{Training algorithm}
\label{sec:algo}


We evaluated all methods on three different architectures: U-Net \citep{ronneberger2015u}, Attention U-Net \citep{oktay2018attention} and Swin-Unet \citep{cao2022swin}.

All models were trained using a stochastic gradient descent optimizer with a variable learning rate and a Nestorov momentum of 0.9 for 500 epochs. The initial learning rate and batch size was selected using a grid search and the model with the best foreground Dice score on the validation set was used for evaluation on the test set. This process was performed separately for each model evaluated.

Data augmentation was applied on the fly during training. The following augmentations were applied: scaling, rotation, Gaussian blur, brightness and contrast adjustment. The data were randomly split into 80\%/10\%/10\% for the training, validation and test sets with the same proportion of images from each dataset in the splits. 

\subsection{Loss functions}
\label{sec:lossfunctions}

In this work we evaluate three different loss functions that have been proposed to deal with the problem of partially labelled data in deep learning-based segmentation and compare them to a baseline CCE loss. The three approaches are representative of the different types of methods that have been proposed in the literature as described in Section \ref{sec:related}. 
A detailed summary of each approach evaluated is given below.

\subsubsection{Categorical Cross Entropy Loss} This is the standard CCE loss that is commonly used when training deep learning segmentation models and we use this as a baseline in our experiments. The loss function is shown in Equation \ref{eq:ce}.
\begin{equation}
\label{eq:ce}
\scriptstyle \mathrm{L}_{CCE} = - \sum_{n \in \Omega _N} y_n \cdot
\log \frac{\exp(\hat{y}_{n})}{\sum_{i \in \Omega _N}\exp(\hat{y}_{i})}
\end{equation}
\noindent where $\hat{y}$ is the network prediction and $y$ is the GT.

\subsubsection{Adaptive CCE loss}
The adaptive CCE (aCCE) loss function was proposed by \cite{mariscal-harana_artificial_2023}. 
The standard CCE loss equation is used as the basis for this loss function. However, for each image and GT segmentation pair, the losses for labels that are not in the GT segmentation are removed from the loss calculation. The adaptive CCE loss function for samples in dataset $k$ is shown in Equation \ref{eq:ace}.
\begin{equation}
\label{eq:ace}
\scriptstyle \mathrm{L}_{aCCE} = - \sum_{n \in \mathcal{L} _k} y_n \cdot
\log \frac{\exp(\hat{y}_{n})}{\sum_{i \in \Omega _N}\exp(\hat{y}_{i})}
\end{equation}

\subsubsection{Marginal loss}
The marginal loss function was proposed by \cite{shi_marginal_2021}.
In this loss function, for each image and label pair, the model predictions for the missing labels are merged with the background prediction before the loss calculation, 
as shown in Equation \ref{eq:qm}.
\begin{equation}
\label{eq:qm}
\scriptstyle \mathrm q_m = \sum_{n\in \Phi _{m}}^{}\frac{exp(\hat{y}_{n})}{\sum_{i \in \Omega _N}\exp(\hat{y}_i)}
\end{equation}
\noindent Here, $q_m$ is the posterior classification probability for merged class $m$ after the missing label(s) and background class probabilities are summed together and $\Phi _{m}$ is the set of classes to be merged to form class $m$. 
Based on this, the marginal loss function is shown in Equation \ref{eq:mce}.
\begin{equation}
\label{eq:mce}
\scriptstyle \mathrm{L }_{marginal} = -\sum_{m \in {\Omega}'_M}^{}z_m \log(q_m)
\end{equation}
\noindent where $z_m$ is the GT one-hot vector, in which the missing labels and background class are now considered to be the same class, and ${\Omega}'_M$ is the new set of class labels after merging. 

\subsubsection{Adaptive BCE Loss}
Finally, the sum of BCE losses was proposed by \cite{petit_handling_2018}, which we term as the adaptive BCE (aBCE) loss in this paper. In this loss function, the loss is only calculated for labels that are present in the GT (not including the background). The loss function is shown in Equation \ref{eq:sbce}.
\begin{equation}
\label{eq:sbce}
\scriptstyle \mathrm{L }_{aBCE} =\sum_{n \in \mathcal{L} _k} -w_n(y_{n}\cdot \log(\hat{y}_{n})+(1-y_{n})\cdot \log(1-\hat{y}_{n})) 
\end{equation}
\noindent where $w_n \in \{0;1\}$ is a binary weight map which ignores pixels for class $n \notin \mathcal{L}_k$ .

\section{Experiments}
A number of experiments were carried out in order to investigate the performance of the loss functions detailed in Section \ref{sec:lossfunctions}. Since the different datasets utilised differ in their image characteristics we also investigate the impact of domain shift on model performance.

First, to illustrate the problem we are trying to address, in Section \ref{sec:expt1} we investigate intra-/inter-domain training/testing using the standard CCE loss function without partial labelling. Following that, a thorough evaluation of the aCCE, marginal and aBCE loss functions is carried out by assessing their performance for training models in intra- (Section \ref{sec:expt2}) and inter-domain (Section \ref{sec:expt3}) settings, with varying levels of fully-labelled and partially-labelled data (with a single missing label). Finally, we examine the performance of the same loss functions when training using partially labelled data with more than one label missing (Section \ref{sec:expt4}).

\subsection{Experiment 1: Intra-/inter-domain training using standard loss function}
\label{sec:expt1}

\begin{figure}[h]
    \centering
    \includegraphics[width=0.8\linewidth]{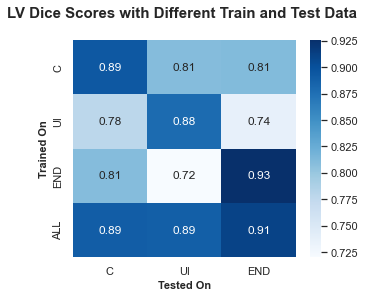}
    \caption{Experiment 1: Test Dice scores achieved by training and evaluating intra-domain and cross-domain LV segmentation models using three different datasets. C = CAMUS \citep{leclerc_deep_2019}, UI = Unity Imaging \citep{huang_fix--step_2023}, END = EchoNet Dynamic \citep{ouyang_video-based_2020}.}
    \label{fig:heatmap}
\end{figure}

The first experiment aims to illustrate the need to train segmentation models with multiple diverse datasets. In this experiment we use fully-labelled data and hence focus on a single class, the LV. We used only the U-Net architecture in this experiment as the problem of domain shift has been established across multiple architectures in the literature \citep{guan2021domain}.
We trained several U-Net segmentation models for LV segmentation using each of the three public datasets mentioned in Section \ref{sec:datasets} individually, and then using all three datasets together. The models were trained with the training algorithm explained in Section \ref{sec:algo} using the standard CCE loss function.

Each model was evaluated on each of the three datasets separately and Figure \ref{fig:heatmap} shows the test set Dice scores achieved on each dataset. As can be seen, the models trained on a single dataset only perform well when tested on the dataset they were trained on. There is a significant drop in performance (of 0.08-0.21 Dice) when testing on data from a different dataset/domain. Training using all the datasets provides robust segmentation performance on all datasets which was comparable to training on each dataset individually.

The conclusion of this experiment is that it is necessary to train models using multiple diverse datasets in order to improve the generalisability of echo segmentation models. 

\subsection{Experiment 2: Intra-domain partial labelling}
\label{sec:expt2}

In this experiment, the loss functions introduced in Section \ref{sec:lossfunctions} were evaluated on a partially-labelled dataset. To remove the domain shift effect between datasets we used only the CAMUS dataset (which contains all three labels) with the LVM artificially removed (i.e. set to background) in 50\% of the GT segmentation masks of the training and validation sets. The LVM label was still available in the test set, enabling quantitative evaluation of the results.

Four models were trained and tested using this modified dataset, one for each of the loss functions: standard CCE loss, aCCE loss, marginal loss and aBCE loss. These models were compared to a benchmark model that was trained using the fully-labelled CAMUS dataset without the artificial removal of labels. As an additional comparison, we also applied the \emph{pseudo-labelling} approach as described in \cite{filbrandt2021learning}. In this approach, predictions for the missing labels were first generated using a model trained using standard CCE with the fully-labelled data (these are called \emph{pseudo-labels}). Subsequently, another model was trained using standard CCE on the combined labels and pseudo-labels, which formed a fully-labelled dataset. This second model was used for inference. All models were trained five times with different random seeds and the average results are reported. These experiments were conducted across three different network architectures: U-Net, Attention U-Net, and Swin-Unet.

\begin{figure*}[p]
    \centering
    \includegraphics[width=6.0in]{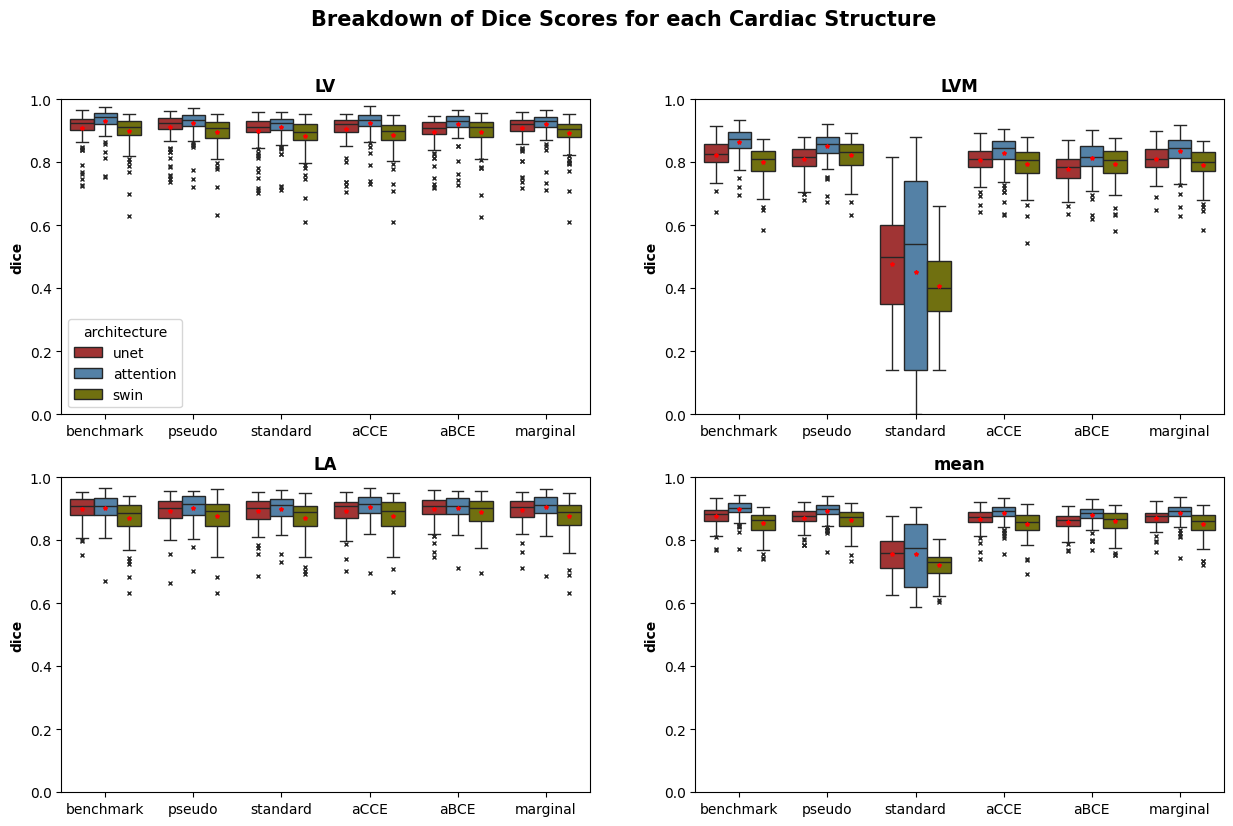} 
    \caption{Experiment 2: Test set results when training using only the CAMUS dataset with 50\% of LVM labels removed. Box-whisker plots show Dice score for each segmented structure and the overall mean. The stars within the box-whisker plots represent the mean.  Red = U-Net, blue = Attention U-Net, green = Swin-Unet.}
    \label{fig:expt2res}
\end{figure*}

\begin{figure*}[p]
\centering
\begin{subfigure}{0.5\linewidth}
  \centering
  \includegraphics[width=0.99\linewidth]{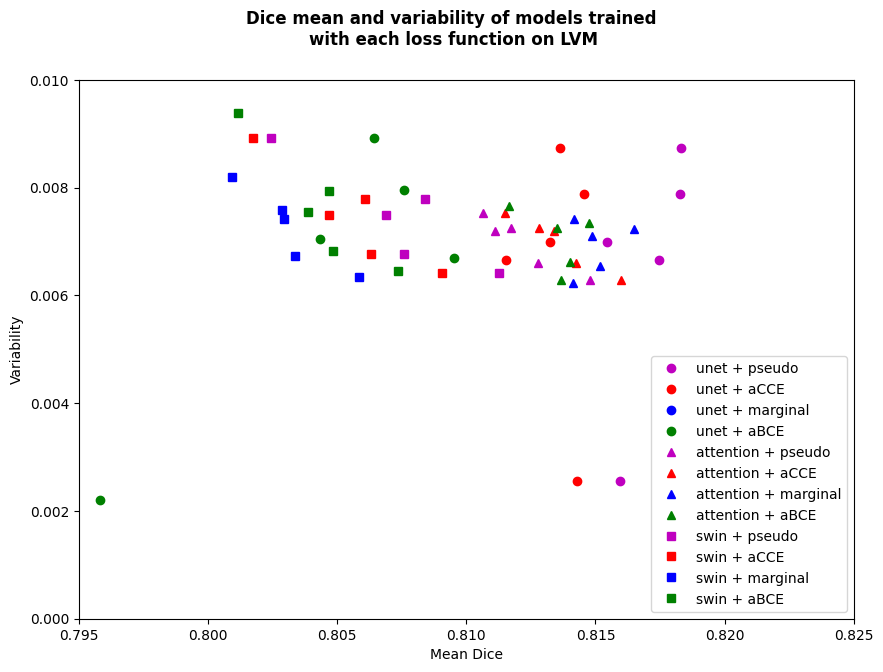}
  \caption{Mean/variability Dice scatter plot}
  \label{fig:expt2scatter}
\end{subfigure}%
\begin{subfigure}{0.5\linewidth}
  \centering
  \includegraphics[width=0.9\linewidth]{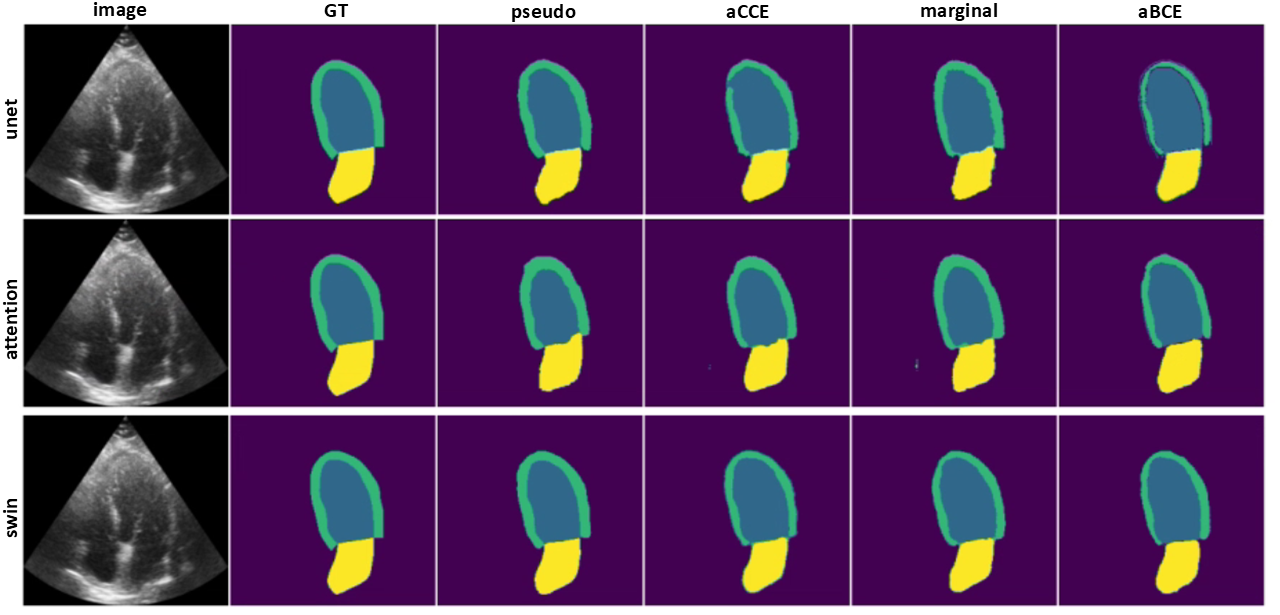}
  \caption{Example predictions}
  \label{fig:expt2preds}
\end{subfigure}
\caption{Experiment 2: Test set results when training using only the CAMUS dataset with 50\% of LVM labels removed. (a) Dice mean and variability (standard deviation) of each loss function and architecture on the LVM. Models were trained five times with different random seeds and the plot shows the mean and standard deviation of each on the test set. (b) Sample predictions from the models. Left to right: image, GT, pseudo-labelling, aCCE, marginal, aBCE.}
\label{fig:expt2}
\end{figure*}

The test set results are shown in Figure \ref{fig:expt2res}. The box-whisker plots show the breakdown of the Dice scores for each class, and show that the pseudo-labelling, aCCE, marginal and aBCE loss models achieved a significantly higher Dice score for the LVM compared to the standard CCE loss model, and performed comparably to the benchmark. On the classes with full labelling (LV and LA) all models performed similarly.
Some sample model predictions and a scatter plot with the Dice mean and variability of the models is shown in Figure \ref{fig:expt2}. In general, similar trends in performance were observed across all three network architectures.

This experiment shows that, in a single domain dataset, all three of the loss functions evaluated are highly effective in dealing with partially labelled data, and perform comparably with a model trained using a fully-labelled dataset. The pseudo-labelling comparative approach also works well. 

\subsection{Experiment 3: Inter-domain partial labelling}
\label{sec:expt3}
\subsubsection{Viability of the loss functions}

In this experiment, we evaluate the viability of the loss functions when there is a domain shift between the differently labelled datasets. For this, we use two datasets, CAMUS and Unity Imaging.

\begin{figure*}[h]
\centering
\begin{subfigure}{0.5\linewidth}
  \centering
  \includegraphics[width=0.9\linewidth]{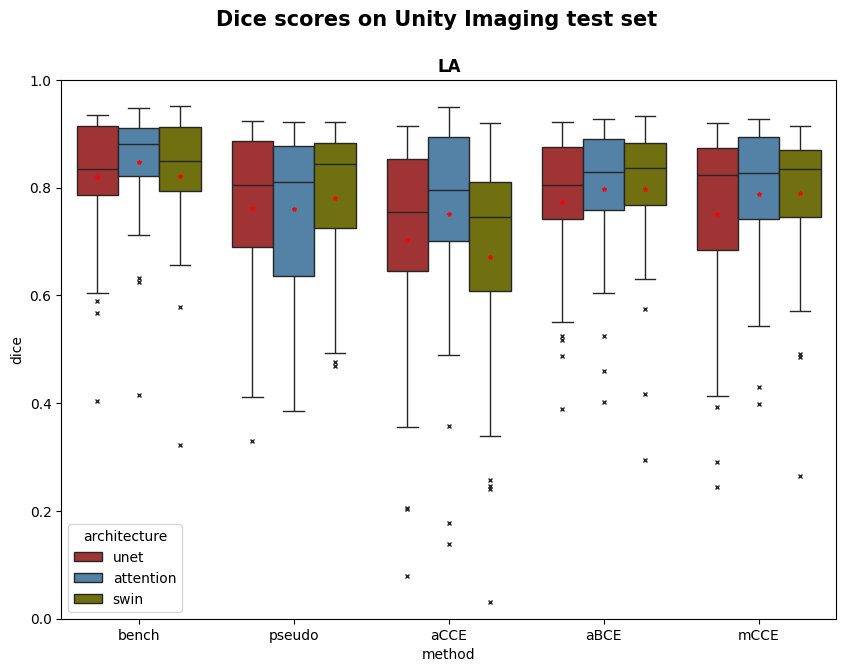}
  \caption{Dice score box-whisker plots for a single run}
  \label{fig:noldboxplot}
\end{subfigure}%
\begin{subfigure}{0.5\linewidth}
  \centering
  \includegraphics[width=0.9\linewidth]{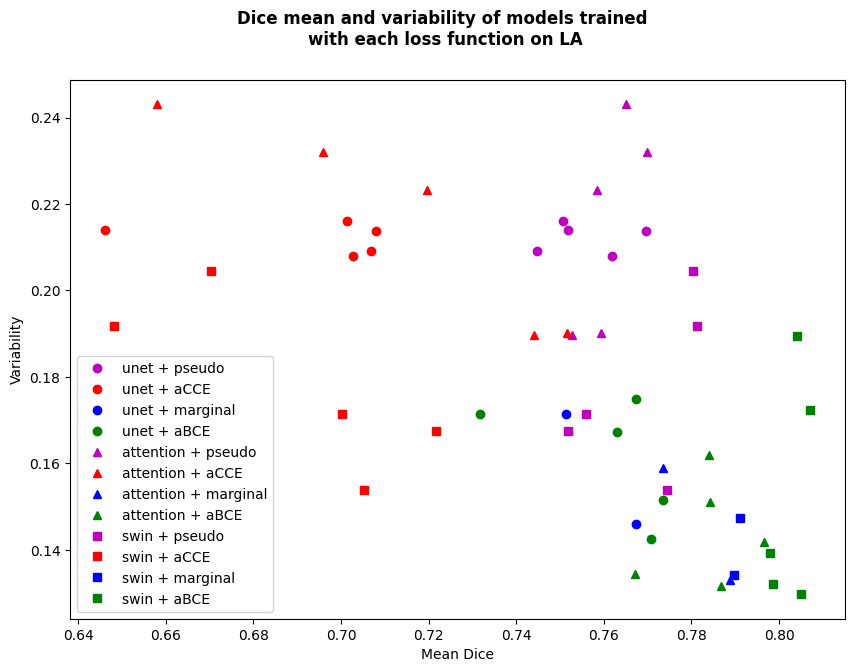}
  \caption{Mean/variability Dice scatter plot}
  \label{fig:noldscatter}
\end{subfigure}
\caption{Experiment 3: Test set results when training using CAMUS and Unity Imaging datasets with the LA removed from Unity Imaging. (a) Box-whisker plots show Dice score for the LA in the Unity Imaging test set. The stars within the box-whisker plots represent the mean. Red = U-Net, blue = Attention U-Net, green = Swin-Unet (b) Scatter plot of Dice mean and variability (i.e. standard deviation) of each loss function and architecture on the LA for images from the Unity Imaging dataset. Models were trained five times with different random seeds and the plot shows the mean and standard deviation of each on the test set.}
\label{fig:expt31}
\end{figure*}

\begin{figure*}[p]
\centering
\begin{subfigure}{.5\textwidth}
  \centering
  \includegraphics[width=1.0\linewidth]{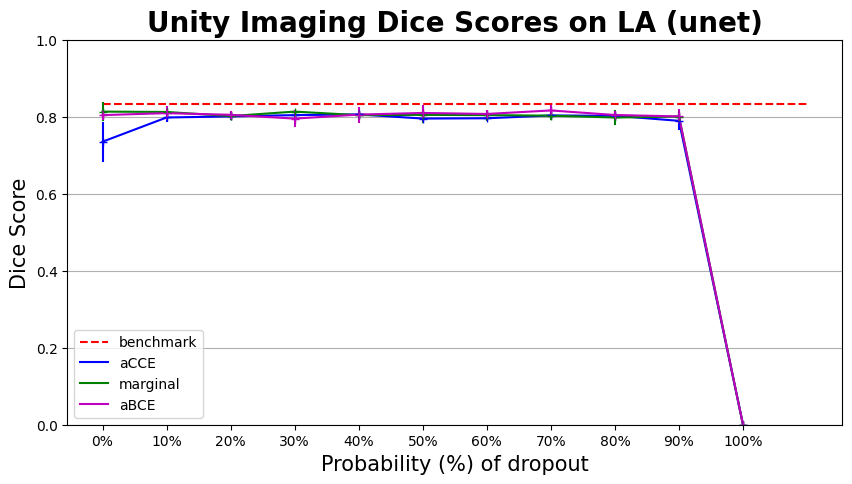}
  \caption{Label dropout graph}
  \label{fig:ldunet}
\end{subfigure}%
\begin{subfigure}{.5\textwidth}
  \centering
  \includegraphics[width=1.0\linewidth]{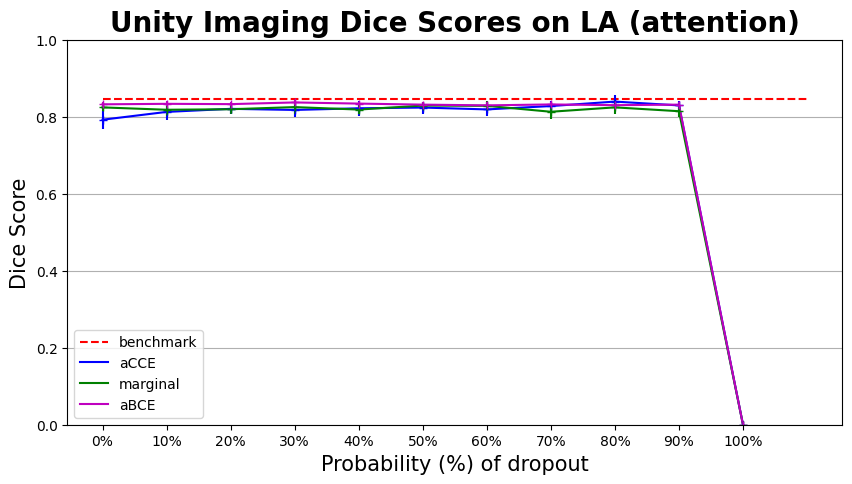}
  \caption{Label dropout graph}
  \label{fig:ldatt}
\end{subfigure}%

\begin{subfigure}{.5\textwidth}
  \centering
  \includegraphics[width=1.0\linewidth]{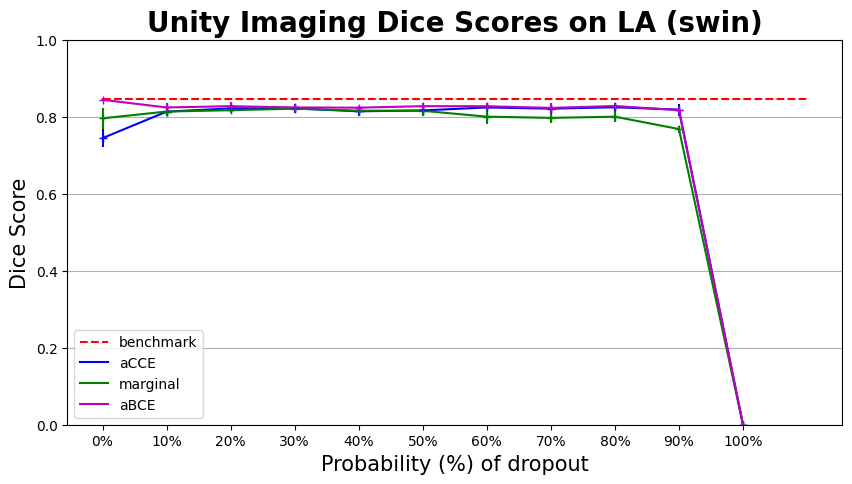}
  \caption{Label dropout graph}
  \label{fig:ldswin}
\end{subfigure}%
\begin{subfigure}{.5\textwidth}
  \centering
  \includegraphics[width=0.8\linewidth]{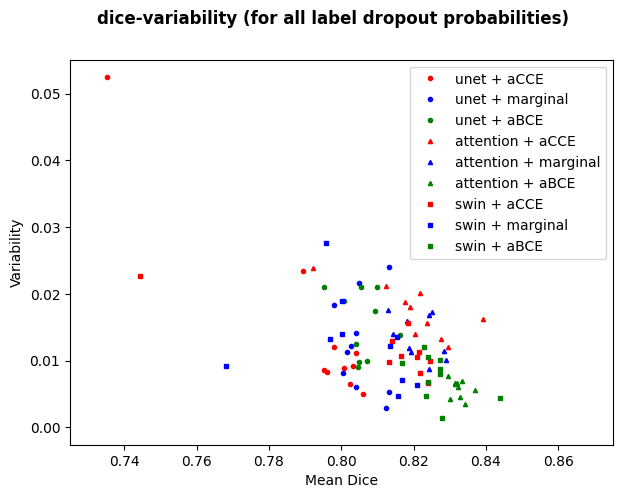}
  \caption{Mean/variability Dice scatter plot}
  \label{fig:ldscatter}
\end{subfigure}
\caption{Experiment 3: Label dropout evaluated when training using CAMUS and Unity Imaging datasets with the LA GT labels removed from the Unity Imaging dataset. (a) Test Dice scores from models trained with different probabilities of label dropout on the LA for images from the Unity Imaging dataset. Models were trained five times with different random seeds and the error bars show the mean and standard deviation of the median Dice scores on each model. The benchmark was trained with a standard CCE loss using a fully labelled dataset. (b) Scatter plot of Dice mean and variability (i.e. standard deviation) of each model trained with the different probabilities of label dropout on the LA for images from the Unity Imaging dataset.}
\label{fig:ldgraph}
\end{figure*}

As shown in Table \ref{tab:datasum}, the CAMUS and Unity Imaging datasets are both fully-labelled. Therefore, in this experiment we artificially remove the LA from all Unity Imaging images in the training and validation sets to simulate a partially labelled dataset. The LA label is still available in the test set for quantitative evaluation. With the resulting partially-labelled dataset, models were trained with the aCCE, marginal and aBCE loss functions. The pseudo-labelling approach was also evaluated, in which a model was first trained on only the fully-labelled CAMUS dataset, which was then used to create the pseudo-labels of the LA in the Unity imaging dataset. The resulting fully-labelled data was used to train the final model. Finally, a benchmark model was trained using fully-labelled data from both CAMUS and Unity datasets. Each model was trained and evaluated five times with different random seeds and the average results reported. This experiment was performed using all three network architectures.

Figure \ref{fig:noldboxplot} shows box-whisker plots for the Dice scores of one run and Figure \ref{fig:noldscatter} shows the relationship between the mean and standard deviation of the Dice scores across the five runs for each approach. It can be seen that, in the presence of domain shift, the marginal and aBCE loss models perform similarly while the aCCE and pseudo-labelling have a lower Dice and higher variability in performance.

We also compared the computational costs of the three loss functions and the standard CCE loss by recording the average training time per epoch. The results showed no significant differences in computational efficiency, with average epoch times varying only within a small range (e.g. variance of $\pm 1.08$ seconds using the U-Net architecture). This trend was consistent across the other network architectures, indicating that the choice of loss function does not substantially affect training time.

\begin{figure*}[p]
\centering
\begin{subfigure}{.5\textwidth}
  \centering
  \includegraphics[width=0.9\linewidth]{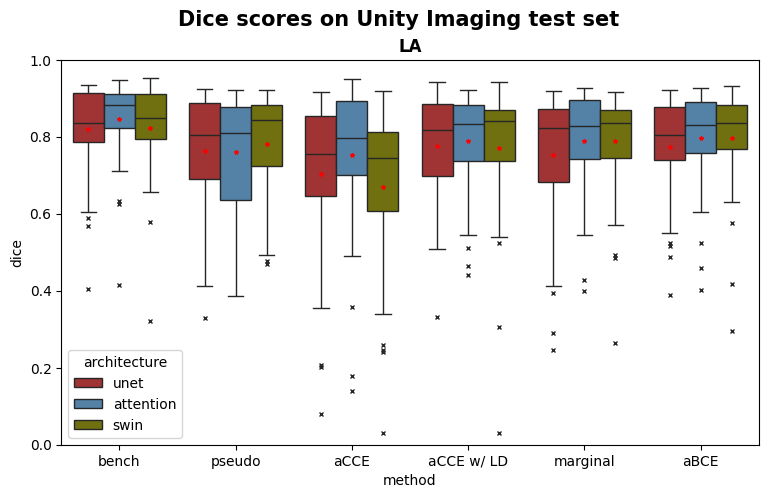}
  \caption{Dice score box-whisker plots}
  \label{fig:ldboxplots}
\end{subfigure}%
\begin{subfigure}{.5\textwidth}
  \centering
  \includegraphics[width=0.9\linewidth]{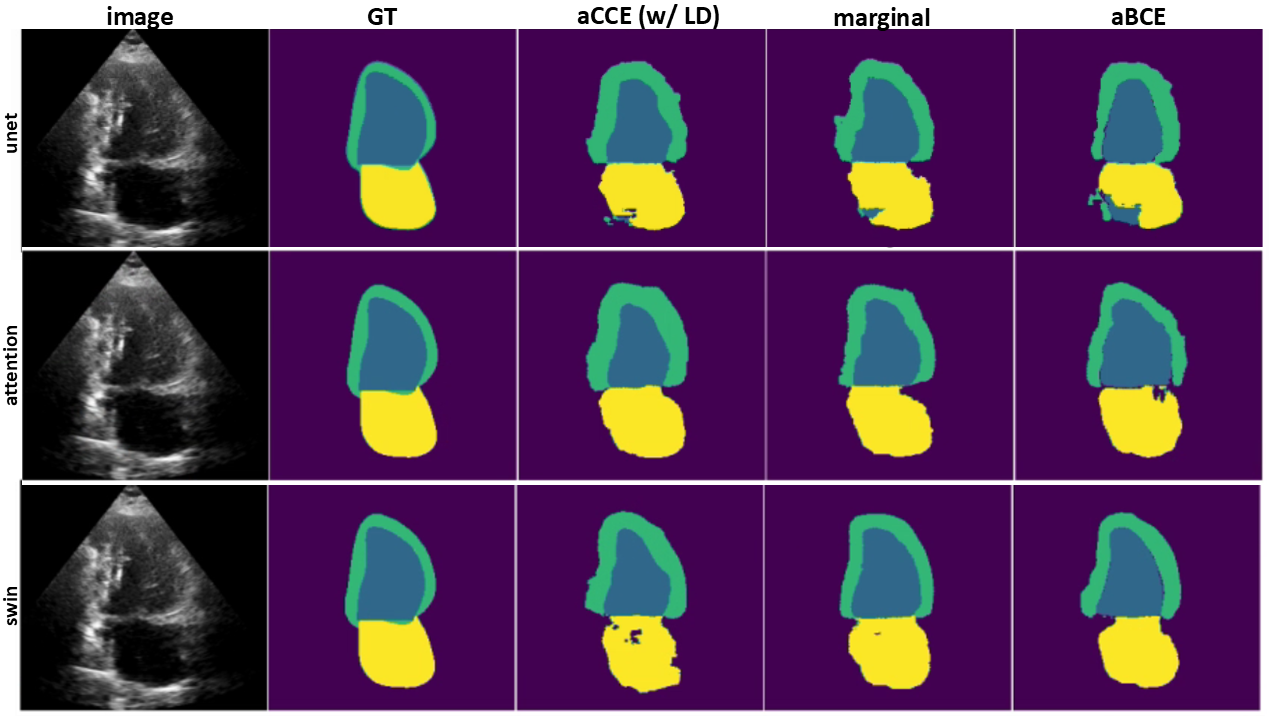}
  \caption{Example predictions}
  \label{fig:ldpreds}
\end{subfigure}
\caption{Experiment 3: Test set results when training using CAMUS and Unity Imaging datasets with the LA GT labels removed from the Unity Imaging dataset. (a) Box-whisker plots show Dice score for the LA in the Unity Imaging test set. The stars within the box-whisker plots represent the mean. Red = U-Net, blue = Attention U-Net, green = Swin-Unet (b) Sample predictions from the models. Left to right: image, GT, aCCE w/ label dropout, marginal, aBCE.}
\label{fig:expt3}
\end{figure*}

In our recent work \citep{mypaper}, it was shown that the aCCE loss performed better in an inter-domain setting when combined with a \emph{label dropout} scheme. Therefore, we also evaluated label dropout on all of the loss functions to see if it improved their performance. The label dropout scheme works by applying a predefined probability that a label will be dropped out, i.e. set to background in the GT segmentations. This dropout happens on the fly during training with each image being considered separately. Only the partially labelled class is dropped out. The motivation of label dropout is to break the link between label presence and domain characteristics.

We evaluate the performance of the three loss functions with different probabilities of label dropout, from 0\% to 100\% at 10\% intervals. The test set Dice scores on the LA in the Unity Imaging images are shown in Figure \ref{fig:ldgraph}. As before, each model was trained 5 times with different random seeds and the results averaged. The benchmark model was trained with the standard CCE loss using a fully-labelled dataset.


Figures \ref{fig:ldunet}, \ref{fig:ldatt} and \ref{fig:ldswin} show that the marginal and aBCE losses perform similarly with and without label dropout across the different network architectures. These loss functions do not seem to suffer a significant drop in performance in the inter-domain setting. The aCCE loss function does exhibit a drop in performance when tested in an inter-domain setting. However, its Dice score improves with label dropout, making its performance comparable with the other two loss functions.

The performance of the models is further analysed in Figure \ref{fig:ldscatter}, which shows how the Dice score and its variability (i.e. the standard deviation of the Dice score) vary with the 10 different label dropout probabilities. As shown in the scatter plot, the aCCE loss models have a slightly higher variability and lower Dice score, with the marginal loss coming in second and the aBCE loss with the lowest variability and highest Dice scores. 

The Dice scores on the Unity Imaging LA for the different models and some sample model predictions are displayed in Figure \ref{fig:expt3}. The box-whisker plot in Figure \ref{fig:ldboxplots} shows a comparison between the benchmark, pseudo-labelling, aCCE with the best performing label dropout from Figures \ref{fig:ldunet}, \ref{fig:ldatt} and \ref{fig:ldswin} (40\% for U-Net, 80\% for Attention U-Net and 30\% for Swin-Unet), marginal, and aBCE models. Here, it is clear that the pseudo-labelling model performs worst and the aCCE model with the label dropout performs comparably to the marginal and aBCE models.

\begin{table*}[p]%
  \centering
  \caption{Systematic evaluation of the impact on performance of the relative number of fully-labelled and partially-labelled data in the training set: CAMUS dataset is fully-labelled and Unity Imaging dataset is partially-labelled (with the LA artificially removed) using the U-Net architecture. The models were trained with the aCCE, marginal and aBCE loss functions and the median Dice scores on the LA in the Unity Imaging test set are displayed for the three network architectures. LA = left atrium.}
  \subfloat[][]{
  \resizebox{0.65\linewidth}{!}{
\begin{tabular}{ccccccc}
\hline
\multicolumn{7}{|c|}{\textbf{Unity Imaging Median Dice scores on LA (U-Net)}}                                                                                                                                                                                                                                                                  \\ \hline
\multicolumn{3}{|c|}{\multirow{2}{*}{}}                                                                                                                                             & \multicolumn{4}{c|}{CAMUS}                                                                                                                     \\ \cline{4-7} 
\multicolumn{3}{|c|}{}                                                                                                                                                              & 100                               & 200                               & 400                               & \multicolumn{1}{c|}{1000}          \\ \hline
\multicolumn{1}{|c|}{\multirow{12}{*}{\rotatebox[origin=c]{90}{\textbf{Unity Imaging (no LA)}}}} & \multicolumn{1}{c|}{\multirow{3}{*}{0}} & \multicolumn{1}{c|}{aCCE}     & \multicolumn{1}{l}{$0.658\pm 0.06$} & \multicolumn{1}{l}{$0.706\pm 0.02$} & \multicolumn{1}{l}{$0.753\pm 0.02$} & \multicolumn{1}{l|}{$0.770\pm 0.01$} \\
\multicolumn{1}{|c|}{}                                                                           & \multicolumn{1}{c|}{}                            & \multicolumn{1}{c|}{marginal} & \multicolumn{1}{l}{$0.661\pm 0.02$} & \multicolumn{1}{l}{$0.696\pm 0.05$} & \multicolumn{1}{l}{$0.746\pm 0.03$} & \multicolumn{1}{l|}{$0.787\pm 0.01$} \\
\multicolumn{1}{|c|}{}                                                                           & \multicolumn{1}{c|}{}                            & \multicolumn{1}{c|}{aBCE}     & \multicolumn{1}{l}{$0.638\pm 0.04$} & \multicolumn{1}{l}{$0.671\pm 0.01$} & \multicolumn{1}{l}{$0.731\pm 0.02$} & \multicolumn{1}{l|}{$0.764\pm 0.04$} \\ \cline{2-7} 
\multicolumn{1}{|c|}{}                                                                           & \multicolumn{1}{c|}{\multirow{3}{*}{100}}        & \multicolumn{1}{c|}{aCCE}     & $0.626\pm 0.06$                     & $0.751\pm 0.02$                     & $0.801\pm 0.01$                     & \multicolumn{1}{c|}{$0.795\pm 0.03$} \\
\multicolumn{1}{|c|}{}                                                                           & \multicolumn{1}{c|}{}                            & \multicolumn{1}{c|}{marginal} & $0.675\pm 0.08$                     & $0.771\pm 0.03$                     & $0.810\pm 0.01$                     & \multicolumn{1}{c|}{$0.808\pm 0.01$} \\
\multicolumn{1}{|c|}{}                                                                           & \multicolumn{1}{c|}{}                            & \multicolumn{1}{c|}{aBCE}     & $0.756\pm 0.01$                     & $0.779\pm 0.05$                     & $0.824\pm 0.02$                     & \multicolumn{1}{c|}{$0.808\pm 0.02$} \\ \cline{2-7} 
\multicolumn{1}{|c|}{}                                                                           & \multicolumn{1}{c|}{\multirow{3}{*}{200}}        & \multicolumn{1}{c|}{aCCE}     & $0.612\pm 0.09$                     & $0.735\pm 0.07$                     & $0.784\pm 0.02$                     & \multicolumn{1}{c|}{$0.799\pm 0.01$} \\
\multicolumn{1}{|c|}{}                                                                           & \multicolumn{1}{c|}{}                            & \multicolumn{1}{c|}{marginal} & $0.756\pm 0.03$                     & $0.790\pm 0.03 $                    & $0.814\pm 0.01$                     & \multicolumn{1}{c|}{$0.817\pm 0.00$} \\
\multicolumn{1}{|c|}{}                                                                           & \multicolumn{1}{c|}{}                            & \multicolumn{1}{c|}{aBCE}     & $0.729\pm 0.0.6 $                   & $0.811\pm 0.01$                     & $0.821\pm 0.01$                     & \multicolumn{1}{c|}{$0.826\pm 0.00$} \\ \cline{2-7} 
\multicolumn{1}{|c|}{}                                                                           & \multicolumn{1}{c|}{\multirow{3}{*}{400}}        & \multicolumn{1}{c|}{aCCE}     & $0.673\pm 0.04$                     & $0.735\pm 0.04$                     & $0.760\pm 0.03$                     & \multicolumn{1}{c|}{$0.804\pm 0.01$} \\
\multicolumn{1}{|c|}{}                                                                           & \multicolumn{1}{c|}{}                            & \multicolumn{1}{c|}{marginal} & $0.747\pm 0.05 $                    & $0.800\pm 0.01$                     & $0.804\pm 0.00$                     & \multicolumn{1}{c|}{$0.821\pm 0.02$} \\
\multicolumn{1}{|c|}{}                                                                           & \multicolumn{1}{c|}{}                            & \multicolumn{1}{c|}{aBCE}     & $0.780\pm 0.03$                     & $0.813\pm 0.02$                     & $0.827\pm 0.02$                     & \multicolumn{1}{c|}{$0.848\pm 0.01$} \\ \hline
\end{tabular}
  }
  \label{tab:unetbal}
  }%
  \qquad
  \subfloat[][]{
  \resizebox{0.65\linewidth}{!}{
\begin{tabular}{ccccccc}
\\ \hline
\multicolumn{7}{|c|}{\textbf{Unity Imaging Median Dice scores on LA (Attention U-Net)}}                                                                                                                                                                                                                                                                  \\ \hline
\multicolumn{3}{|c|}{\multirow{2}{*}{}}                                                                                                                                             & \multicolumn{4}{c|}{CAMUS}                                                                                                                     \\ \cline{4-7} 
\multicolumn{3}{|c|}{}                                                                                                                                                              & 100                               & 200                               & 400                               & \multicolumn{1}{c|}{1000}          \\ \hline
\multicolumn{1}{|c|}{\multirow{12}{*}{\rotatebox[origin=c]{90}{\textbf{Unity Imaging (no LA)}}}} & \multicolumn{1}{c|}{\multirow{3}{*}{0}} 
& \multicolumn{1}{c|}{aCCE}     & \multicolumn{1}{l}{$0.702\pm 0.01$} & \multicolumn{1}{l}{$0.706\pm 0.02$} & \multicolumn{1}{l}{$0.779\pm 0.01$} & \multicolumn{1}{l|}{$0.810\pm 0.01$} \\
\multicolumn{1}{|c|}{}                                                                           & \multicolumn{1}{c|}{}                            
& \multicolumn{1}{c|}{marginal} & \multicolumn{1}{l}{$0.726\pm 0.00$} & \multicolumn{1}{l}{$0.770\pm 0.01$} & \multicolumn{1}{l}{$0.795\pm 0.01$} & \multicolumn{1}{l|}{$0.801\pm 0.02$} \\
\multicolumn{1}{|c|}{}                                                                           & \multicolumn{1}{c|}{}                            
& \multicolumn{1}{c|}{aBCE}     & \multicolumn{1}{l}{$0.704\pm 0.01$} & \multicolumn{1}{l}{$0.757\pm 0.03$} & \multicolumn{1}{l}{$0.784\pm 0.01$} & \multicolumn{1}{l|}{$0.805\pm 0.01$} \\ \cline{2-7} 
\multicolumn{1}{|c|}{}                                                                           & \multicolumn{1}{c|}{\multirow{3}{*}{100}}        
& \multicolumn{1}{c|}{aCCE}     & $0.699\pm 0.01$                     & $0.794\pm 0.02$                     & $0.801\pm 0.01$                     & \multicolumn{1}{c|}{$0.817\pm 0.01$} \\
\multicolumn{1}{|c|}{}                                                                           & \multicolumn{1}{c|}{}                            
& \multicolumn{1}{c|}{marginal} & $0.794\pm 0.01$                     & $0.820\pm 0.01$                     & $0.814\pm 0.01$                     & \multicolumn{1}{c|}{$0.811\pm 0.02$} \\
\multicolumn{1}{|c|}{}                                                                           & \multicolumn{1}{c|}{}                            
& \multicolumn{1}{c|}{aBCE}     & $0.754\pm 0.00$                     & $0.819\pm 0.01$                     & $0.821\pm 0.00$                     & \multicolumn{1}{c|}{$0.826\pm 0.01$} \\ \cline{2-7} 
\multicolumn{1}{|c|}{}                                                                           & \multicolumn{1}{c|}{\multirow{3}{*}{200}}        
& \multicolumn{1}{c|}{aCCE}     & $0.702\pm 0.01$                     & $0.767\pm 0.03$                     & $0.798\pm 0.02$                     & \multicolumn{1}{c|}{$0.809\pm 0.02$} \\
\multicolumn{1}{|c|}{}                                                                           & \multicolumn{1}{c|}{}                            
& \multicolumn{1}{c|}{marginal} & $0.774\pm 0.02$                     & $0.816\pm 0.02$                     & $0.812\pm 0.01$                     & \multicolumn{1}{c|}{$0.816\pm 0.01$} \\
\multicolumn{1}{|c|}{}                                                                           & \multicolumn{1}{c|}{}                            
& \multicolumn{1}{c|}{aBCE}     & $0.732\pm 0.03$                     & $0.799\pm 0.01$                     & $0.808\pm 0.01$                     & \multicolumn{1}{c|}{$0.829\pm 0.01$} \\ \cline{2-7} 
\multicolumn{1}{|c|}{}                                                                           & \multicolumn{1}{c|}{\multirow{3}{*}{400}}        
& \multicolumn{1}{c|}{aCCE}     & $0.681\pm 0.06$                     & $0.771\pm 0.02$                     & $0.796\pm 0.02$                     & \multicolumn{1}{c|}{$0.802\pm 0.02$} \\
\multicolumn{1}{|c|}{}                                                                           & \multicolumn{1}{c|}{}                            
& \multicolumn{1}{c|}{marginal} & $0.783\pm 0.01$                     & $0.813\pm 0.02$                     & $0.812\pm 0.00$                     & \multicolumn{1}{c|}{$0.835\pm 0.01$} \\
\multicolumn{1}{|c|}{}                                                                           & \multicolumn{1}{c|}{}                            
& \multicolumn{1}{c|}{aBCE}     & $0.746\pm 0.02$                     & $0.827\pm 0.02$                     & $0.789\pm 0.00$                     & \multicolumn{1}{c|}{$0.830\pm 0.01$} \\ \hline
\end{tabular}
  }
  \label{tab:attbal}
  }
  \qquad
  \subfloat[][]{
  \resizebox{0.65\linewidth}{!}{
\begin{tabular}{ccccccc}
\hline
\multicolumn{7}{|c|}{\textbf{Unity Imaging Median Dice scores on LA (Swin-Unet)}}                                                                                                                                                                                                                                                                  \\ \hline
\multicolumn{3}{|c|}{\multirow{2}{*}{}}                                                                                                                                             & \multicolumn{4}{c|}{CAMUS}                                                                                                                     \\ \cline{4-7} 
\multicolumn{3}{|c|}{}                                                                                                                                                              & 100                               & 200                               & 400                               & \multicolumn{1}{c|}{1000}          \\ \hline
\multicolumn{1}{|c|}{\multirow{12}{*}{\rotatebox[origin=c]{90}{\textbf{Unity Imaging (no LA)}}}} & \multicolumn{1}{c|}{\multirow{3}{*}{0}} 
& \multicolumn{1}{c|}{aCCE}     & \multicolumn{1}{l}{$0.754\pm 0.01$} & \multicolumn{1}{l}{$0.805\pm 0.02$} & \multicolumn{1}{l}{$0.795\pm 0.01$} & \multicolumn{1}{l|}{$0.831\pm 0.01$} \\
\multicolumn{1}{|c|}{}                                                                           & \multicolumn{1}{c|}{}                            
& \multicolumn{1}{c|}{marginal} & \multicolumn{1}{l}{$0.770\pm 0.01$} & \multicolumn{1}{l}{$0.818\pm 0.01$} & \multicolumn{1}{l}{$0.815\pm 0.01$} & \multicolumn{1}{l|}{$0.824\pm 0.00$} \\
\multicolumn{1}{|c|}{}                                                                           & \multicolumn{1}{c|}{}                            
& \multicolumn{1}{c|}{aBCE}     & \multicolumn{1}{l}{$0.772\pm 0.01$} & \multicolumn{1}{l}{$0.806\pm 0.01$} & \multicolumn{1}{l}{$0.806\pm 0.00$} & \multicolumn{1}{l|}{$0.834\pm 0.01$} \\ \cline{2-7} 
\multicolumn{1}{|c|}{}                                                                           & \multicolumn{1}{c|}{\multirow{3}{*}{100}}        
& \multicolumn{1}{c|}{aCCE}     & $0.609\pm 0.07$                     & $0.800\pm 0.01$                     & $0.814\pm 0.02$                     & \multicolumn{1}{c|}{$0.834\pm 0.01$} \\
\multicolumn{1}{|c|}{}                                                                           & \multicolumn{1}{c|}{}                            
& \multicolumn{1}{c|}{marginal} & $0.702\pm 0.04$                     & $0.781\pm 0.02$                     & $0.809\pm 0.01$                     & \multicolumn{1}{c|}{$0.831\pm 0.01$} \\
\multicolumn{1}{|c|}{}                                                                           & \multicolumn{1}{c|}{}                            
& \multicolumn{1}{c|}{aBCE}     & $0.780\pm 0.01$                     & $0.827\pm 0.01$                     & $0.832\pm 0.01$                     & \multicolumn{1}{c|}{$0.830\pm 0.00$} \\ \cline{2-7} 
\multicolumn{1}{|c|}{}                                                                           & \multicolumn{1}{c|}{\multirow{3}{*}{200}}        
& \multicolumn{1}{c|}{aCCE}     & $0.464\pm 0.04$                     & $0.761\pm 0.02$                     & $0.783\pm 0.03$                     & \multicolumn{1}{c|}{$0.822\pm 0.01$} \\
\multicolumn{1}{|c|}{}                                                                           & \multicolumn{1}{c|}{}                            
& \multicolumn{1}{c|}{marginal} & $0.598\pm 0.07$                     & $0.748\pm 0.05$                     & $0.821\pm 0.01$                     & \multicolumn{1}{c|}{$0.826\pm 0.01$} \\
\multicolumn{1}{|c|}{}                                                                           & \multicolumn{1}{c|}{}                            
& \multicolumn{1}{c|}{aBCE}     & $0.774\pm 0.03$                     & $0.831\pm 0.00$                     & $0.835\pm 0.01$                     & \multicolumn{1}{c|}{$0.834\pm 0.01$} \\ \cline{2-7} 
\multicolumn{1}{|c|}{}                                                                           & \multicolumn{1}{c|}{\multirow{3}{*}{400}}        
& \multicolumn{1}{c|}{aCCE}     & $0.519\pm 0.07$                     & $0.748\pm 0.02$                     & $0.789\pm 0.03$                     & \multicolumn{1}{c|}{$0.833\pm 0.01$} \\
\multicolumn{1}{|c|}{}                                                                           & \multicolumn{1}{c|}{}                            
& \multicolumn{1}{c|}{marginal} & $0.605\pm 0.04$                     & $0.733\pm 0.03$                     & $0.807\pm 0.02$                     & \multicolumn{1}{c|}{$0.829\pm 0.01$} \\
\multicolumn{1}{|c|}{}                                                                           & \multicolumn{1}{c|}{}                            
& \multicolumn{1}{c|}{aBCE}     & $0.769\pm 0.01$                     & $0.820\pm 0.01$                     & $0.834\pm 0.02$                     & \multicolumn{1}{c|}{$0.839\pm 0.01$} \\ \hline
\end{tabular}
  }
  \label{tab:swinbal}
  }%
  \label{tab:bal}%
\end{table*}

\subsubsection{Annotation consistency study}
\label{sec:expt3_anndiff}

We also assessed the consistency of annotations between the CAMUS and Unity Imaging datasets in order to evaluate whether differences in annotation protocol affect performance. We performed an experiment in which a subset of 10 images from the CAMUS dataset (the sources) were registered to similar images from the Unity Imaging dataset (the targets), and the resulting transformations were applied to the corresponding source GT labels. The Dice similarity coefficient between the transformed source labels and the target labels yielded an average of 0.83 ± 0.07, indicating a moderate level of annotation discrepancy across datasets. This variation is expected given differences in labelling protocols and image characteristics. Nonetheless, it does not affect our main conclusion that the proposed loss functions enable the model to successfully predict all three anatomical structures, rather than missing them due to inter-dataset domain shifts.

\subsubsection{Impact of relative numbers of fully-labelled and partially-labelled data}
\label{sec:sys}

Next, the relationship between the relative numbers of partially-labelled images and fully-labelled images was investigated. The previous experimental setup was repeated, but this time the number of images from Unity imaging and CAMUS were varied. These models were trained three times each and the results averaged. This experiment was performed using all three network architectures.

The test set median Dice scores on the LA in the Unity imaging images are shown in Tables 2a, 2b and 2c.
As can be seen, training on only the CAMUS dataset (which includes 1000 fully-labelled images) yields similar performance to training with 200 fully-labelled images (from CAMUS) and 100 partially-labelled images (from Unity Imaging) using the U-Net and Attention U-Net architectures. This suggests that the model is able to learn features of the missing label in the partially-labelled dataset. The addition of extra (partially labelled) data from the existing domain has a more beneficial effect on model performance than extra (fully labelled) data from the a different domain.
Additionally, the models yield similar Dice scores when training with 400 or more CAMUS and 200 or more Unity Imaging images across all three architectures. This indicates that, after a certain number of training images is reached, further increases in the training set size lead to only small increases in performance. However, this `plateauing' point is reached with a smaller number of images when the data come from different domains, even if they are partially labelled. Therefore, training with multiple diverse datasets, even if partially-labelled, can reduce the data demands compared to training with a single fully-labelled dataset. Notably, the best performance for all loss functions was achieved when training with all available (partially labelled) data.

\subsection{Experiment 4: Inter-domain partial labelling (all datasets)}
\label{sec:expt4}

The best methods from the previous experiment were then taken forward to train models with all three datasets mentioned in Section \ref{sec:datasets}. Recall that the CAMUS and Unity Imaging datasets are fully-labelled while the EchoNet-Dynamic dataset is partially-labelled.
Three models were trained: the aCCE loss model with label dropout (the best performance from Figure \ref{fig:ldgraph}: 40\% for U-Net, 80\% for Attention U-Net and 30\% for Swin-Unet), the marginal loss model and the aBCE loss model (both without label dropout as this did not improve their performance). This experiment was performed using all three network architectures.

Because of the much larger size of EchoNet-Dynamic compared to the other datasets, in addition to training with all available data we also trained models using a dataset with an approximately equal number of partially- and fully-labelled images (1400 from EchoNet-Dynamic, 1000 from CAMUS and 400 from Unity Imaging).

As the EchoNet-Dynamic GT segmentations do not include labels for the LVM and LA, a trained observer annotated these structures for 20 images from the EchoNet-Dynamic test set. These manual annotations were then checked by a cardiologist. These extra GT segmentations were used to produce quantitative results for the experiment and the results are shown in Table \ref{tab:finalres}. It can be seen that the marginal loss achieves the highest mean Dice score out of the three loss functions for the U-Net and Attention U-Net architectures, slightly outperforming the aBCE loss in second place whereas for the Swin-Unet, the aBCE loss slightly outperforms the marginal loss. \revision{The mean absolute error (MAE) between the ground-truth EF and EF calculated using the model predicted segmentations on models trained using U-Net and Attention U-Net architectures remain low when using the marginal loss and aBCE loss, whereas models trained using Swin-Unet produce a low MAE only when using the aBCE loss.}
The models trained with the balanced dataset do not have a large difference in Dice scores compared to those trained with the full dataset. In fact, for some structures the Dice scores are lower. This shows that, for the loss functions evaluated, including a large amount of partially-labelled data does not have an adverse affect on the performance of the models. 

\begin{table*}[h]
\centering
\caption{\revision{Experiment 4: Mean test Dice scores over 20 random images from EchoNet Dynamic when trained on EchoNet Dynamic, CAMUS and Unity Imaging data. Dice scores are displayed for each of the three cardiac structures segmented and their overall mean using the three network architectures. Additionally, the MAE between the ground truth EF and EF calculated from model predictions are displayed. LV = left ventricle, LVM = left ventricular myocardium, LA = left atrium, EF = ejection fraction, MAE = mean absolute error.}}
\label{tab:finalres}
\resizebox{0.62\linewidth}{!}{
\begin{tabular}{|c|l|l|l|l|c|c|}
\hline
\multicolumn{1}{|l|}{} &
  \multicolumn{1}{c|}{\textbf{Loss function}} &
  \multicolumn{1}{c|}{\textbf{LV}} &
  \multicolumn{1}{c|}{\textbf{LVM}} &
  \multicolumn{1}{c|}{\textbf{LA}} &
  \textbf{Mean} &
  \textbf{\begin{tabular}[c]{@{}c@{}}EF MAE \\ (\%)\end{tabular}} \\ \hline
\multirow{6}{*}{\textbf{U-Net}}     & \textit{aCCE loss}                & 0.910 & 0.518 & 0.689 & 0.741          & 3.96           \\
                                    & \textit{Marginal loss}            & 0.901 & 0.701 & 0.693 & \textbf{0.819} & \textbf{1.02}  \\
                                    & \textit{aBCE loss}                & 0.908 & 0.635 & 0.731 & 0.809          & 7.11           \\ \cline{2-7} 
                                    & \textit{aCCE loss (balanced)}     & 0.860 & 0.242 & 0.620 & 0.574          & 23.7           \\
                                    & \textit{Marginal loss (balanced)} & 0.882 & 0.668 & 0.675 & \textbf{0.742} & 2.61           \\
                                    & \textit{aBCE loss (balanced)}     & 0.881 & 0.578 & 0.690 & 0.716          & \textbf{2.24}  \\ \hline
\multirow{6}{*}{\textbf{\begin{tabular}[c]{@{}c@{}}Attention \\ U-Net\end{tabular}}} &
  \textit{aCCE loss} &
  0.880 &
  0.421 &
  0.524 &
  0.608 &
  15.8 \\
                                    & \textit{Marginal loss}            & 0.878 & 0.695 & 0.669 & \textbf{0.747} & 1.80           \\
                                    & \textit{aBCE loss}                & 0.885 & 0.622 & 0.692 & 0.733          & \textbf{0.447} \\ \cline{2-7} 
                                    & \textit{aCCE loss (balanced)}     & 0.864 & 0.385 & 0.572 & 0.607          & 0.867          \\
                                    & \textit{Marginal loss (balanced)} & 0.866 & 0.651 & 0.603 & \textbf{0.707} & \textbf{2.17}  \\
                                    & \textit{aBCE loss (balanced)}     & 0.881 & 0.495 & 0.689 & 0.689          & 2.52           \\ \hline
\multirow{6}{*}{\textbf{Swin-Unet}} & \textit{aCCE loss}                & 0.870 & 0.374 & 0.579 & 0.608          & 4.00           \\
                                    & \textit{Marginal loss}            & 0.880 & 0.654 & 0.760 & 0.764          & 4.19           \\
                                    & \textit{aBCE loss}                & 0.884 & 0.671 & 0.757 & \textbf{0.771} & \textbf{1.67}  \\ \cline{2-7} 
                                    & \textit{aCCE loss (balanced)}     & 0.868 & 0.230 & 0.743 & 0.614          & 5.49           \\
                                    & \textit{Marginal loss (balanced)} & 0.873 & 0.655 & 0.750 & 0.760          & 3.13           \\
                                    & \textit{aBCE loss (balanced)}     & 0.886 & 0.687 & 0.765 & \textbf{0.779} & \textbf{0.865} \\ \hline
\end{tabular}%
}
\end{table*}

\begin{figure*}[hbt!]
    \centering
    \includegraphics[width=6in]{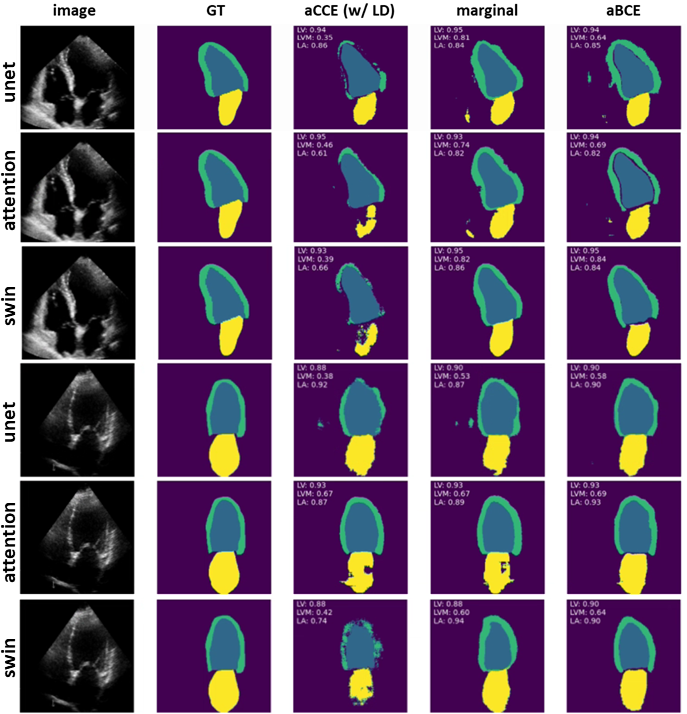} 
    \caption{Experiment 4: Sample predictions produced by the models of the three different loss functions and architectures using images from the annotated EchoNet-Dynamic test set and the external validation clinical dataset. The top three rows are images from the EchoNet-Dynamic dataset and the bottom three rows and images from the external validation clinical dataset.}
    \label{fig:finalrespreds}
\end{figure*}

Across the three loss functions, LV and LA have similar Dice scores, whereas there are more noticeable differences for the LVM. 
The marginal loss performed better than the aBCE loss (Wilcoxon Signed Rank test, p\textless 0.05) and the aBCE loss performed better than the aCCE loss (Wilcoxon Signed Rank test, p\textless0.05) on the LVM for the U-Net and Attention U-Net architectures.
The differences in the model performances on the LV and LA were not found to be statistically significant across the different loss functions and architectures. This is likely due to (i) the LV being labelled in all images in the dataset and (ii) the LA being easier to segment than the LVM due to the shape and contrast at the boundaries of the structure. All loss functions achieved their lowest Dice scores on the LVM with the aCCE performing worse than the others. Sample predictions from the models are shown in Figure \ref{fig:finalrespreds}. These results will be further discussed in Section \ref{sec:dis}.

\begin{table}[]
\centering
\caption{Experiment 4: Mean test Dice scores over 20 images from the GSTFT clinical database when trained on EchoNet Dynamic, CAMUS and Unity Imaging data. Dice scores are displayed for each of the three cardiac structures segmented and their overall mean using the three network architectures. The best mean Dice for each architecture is shown in bold. LV = left ventricle, LVM = left ventricular myocardium, LA = left atrium.}
\label{tab:gsttres}
\resizebox{\linewidth}{!}{
\begin{tabular}{|c|l|l|l|l|l|}
\hline
\multicolumn{1}{|l|}{} &
  \multicolumn{1}{c|}{\textbf{Loss function}} &
  \multicolumn{1}{c|}{\textbf{LV}} &
  \multicolumn{1}{c|}{\textbf{LVM}} &
  \multicolumn{1}{c|}{\textbf{LA}} &
  \multicolumn{1}{c|}{\textbf{Mean}} \\ \hline
\multirow{3}{*}{\textbf{U-Net}}     & \textit{aCCE loss}     & 0.863 & 0.419 & 0.731 & 0.671 \\
                                    & \textit{Marginal loss} & 0.884 & 0.500 & 0.700 & 0.695 \\
                                    & \textit{aBCE loss}     & 0.895 & 0.541 & 0.732 & \textbf{0.723} \\ \hline
\multirow{3}{*}{\textbf{\begin{tabular}[c]{@{}c@{}}Attention\\ U-Net\end{tabular}}} &
  \textit{aCCE loss} &
  0.904 &
  0.579 &
  0.755 &
  0.746 \\
                                    & \textit{Marginal loss} & 0.904 & 0.584 & 0.759 & \textbf{0.749} \\
                                    & \textit{aBCE loss}     & 0.904 & 0.577 & 0.752 & 0.745 \\ \hline
\multirow{3}{*}{\textbf{Swin-Unet}} & \textit{aCCE loss}     & 0.891 & 0.540 & 0.682 & 0.704 \\
                                    & \textit{Marginal loss} & 0.909 & 0.607 & 0.771 & 0.762 \\
                                    & \textit{aBCE loss}     & 0.908 & 0.615 & 0.770 & \textbf{0.764} \\ \hline
\end{tabular}
}
\end{table}

An external validation was also carried out using 20 images from the GSTFT database detailed in Section \ref{sec:data}, to assess the generalisability of the models, and the results are shown in Table \ref{tab:gsttres}. The external validation results reflect a similar trend to the internal evaluation, with the aBCE loss performing best for the U-Net, the marginal loss achieving the highest performance for the Attention U-Net, and both the marginal and aBCE losses yielding comparable results for the Swin-Unet. \revision{These results provide an initial indication of cross-domain generalisation, though broader validation across larger and more heterogeneous cohorts would be valuable to fully characterise robustness.}

\section{Discussion}
\label{sec:dis}
In this paper, we have performed a thorough analysis of three representative loss functions for dealing with the problem of partial labelling in the presence of a domain shift in deep learning-based segmentation. This is the first time that such a quantitative comparative analysis has been performed in any segmentation application with multiple diverse datasets and (with the exception of our recent initial work, \cite{mypaper}) the first time that partial labelling has been investigated for echo segmentation. Through a series of experiments we have shown the need for training with multiple diverse datasets in echo segmentation, compared the performances of the three loss functions in intra- and inter-domain settings, and examined the impact of relative dataset sizes between fully- and partially-labelled data.

Specifically, Experiment 1 demonstrated the need for training echo segmentation models using multiple diverse datasets for robust performance across all domains.

Practically speaking, the use of multiple datasets in multi-structure echo segmentation necessitates the use of partially-labelled data. Therefore, in Experiment 2, we evaluated three loss functions that have been proposed for dealing with partially labelled data. It was shown that, in the intra-domain setting, the aCCE, marginal and aBCE losses all effectively dealt with the partial labels and robust performance was achieved for all structures.

In Experiment 3, we saw that models trained using the aCCE loss function experienced a drop in performance when used in the inter-domain setting, whereas models trained with the marginal and aBCE losses do not. The previously proposed label dropout technique \citep{mypaper} reduced this performance gap but still it was the case that the marginal and aBCE loss functions achieved the most robust performance.
Experiment 3 also investigated the impact of the relative numbers of fully- and partially-labelled data on the loss functions. For all loss functions, it was found that including data from different domains improved performance and reduced the need for large amounts of training data.

Finally, in Experiment 4 we evaluated the loss functions on the realistic scenario of using all available datasets. Robust performance was observed, with the marginal loss function achieving the best quantitative results.

The differences in performance of the three loss functions are likely due to their mathematical formulations and the characteristics of the structures being segmented. Below we discuss a number of differences between the loss formulations and attempt to link them to examples seen in our results.

\subsection{Analysis of loss functions}

\paragraph{Does the loss consider predictions of missing labels?}
The formulations of the aCCE and the aBCE loss functions do not include the missing labels in the loss calculation and therefore the predicted segmentations for these labels in the partially-labelled samples are not considered during training. The ability of models trained using these losses to predict the partial labels is based only on the features learnt from the fully-labelled data. In contrast, the marginal loss does include these predictions in the loss calculation and is considered to be correct if it is either background or the missing class. Intuitively, this would suggest that the marginal loss would achieve more robust performance on the partially labelled class(es), and such differences were observed in Figure \ref{fig:finalrespreds}, in which noticeably better performance was seen on the LVM for the marginal loss model. Differences were not so noticeable on the LA but this is an easier structure to segment.


\paragraph{Does the loss enforce mutual exclusivity during training?}
The aBCE loss treats each class as a binary classification problem (i.e. foreground vs. background). The Sigmoid activation function is used during training, which does not assume mutually exclusivity of the classes (unlike the Softmax employed during training of the aCCE and marginal losses). Treating each class independently in this way  could lead to problems at the boundaries between classes as a pixel near the boundary is more likely to be predicted by the model as belonging to more than one class. This could cause uncertainty as to which class the pixel belongs to. For example, the LV and LVM share a very long boundary. In Figure \ref{fig:finalrespreds} that the aBCE loss model predictions feature a gap between the LV and LVM along this boundary, and this is likely due to the use of the Sigmoid function during training, as these boundary pixels may have been penalised heavily. Note that the aBCE loss uses the Softmax during inference so it does enforce mutual exclusivity, but performance at boundaries likely suffers due to the mismatch between training and inference activation functions.

\paragraph{Failure mode analysis:}
Upon qualitative inspection, we observed that segmentation errors were most evident in the LVM. Unlike the LV and LA, which are more compact and have fewer boundary pixels, the LVM has a longer and more irregular perimeter that varies considerably across subjects and cardiac phases. In addition, the boundaries between the myocardium and adjacent regions are often low in contrast, making them difficult to delineate even in the GT. These factors, combined with inter-dataset domain shifts, increase the likelihood of inconsistent predictions in this region.

The aCCE loss struggled the most with the LVM, producing patchy and discontinuous segmentations. This is likely because the adaptive weighting focuses on handling missing or uncertain labels rather than enforcing spatial or structural consistency, leading to fragmented predictions in regions with complex boundaries. The aBCE loss produced more coherent results overall, but in several cases a small gap appeared between the LVM and LV regions. This gap may be due to the loss function treating each class independently, which can weaken the spatial relationship between adjacent structures. In contrast, the marginal loss achieved the most consistent LVM delineation, particularly under multiple partial-label scenarios, as it encourages balanced supervision across available labels and implicitly regularizes inter-class boundaries.

The LA was segmented well across all three loss functions even though, like the LVM, it was a missing label in the training data. This is likely because it is a clearly defined structure with high contrast and limited shape variability. 

Overall, these observations suggest that differences in structural complexity and boundary definition play a key role in determining where each loss function succeeds or fails.

\subsection{Analysis of pseudo-labelling model}
The pseudo-labelling model has been seen to perform as well as the partial labelling loss functions in Experiment 2 (Section \ref{sec:expt2}) in the intra-domain setting. However, in the inter-domain setting in Experiment 3 (Section \ref{sec:expt3}), the performance dropped significantly. There could be a number of reasons for this, but the main reason is likely due to the image and GT segmentation based domain shift between the datasets. In Experiment 3, as the model is first trained on only the CAMUS dataset, it is not exposed to the domain characteristics of the Unity Imaging dataset before the pseudo-labels are generated. There are differences between these datasets in the contrast and brightness of the cardiac structures which could produce errors in the generation of the pseudo-labels. Furthermore, there are also some differences in annotation style between the two datasets, which could mean the GT segmentations of the LA from the Unity Imaging dataset do not fully match the pseudo-labels.

\subsection{Limitations}
While this study provides valuable insights into the topic of loss functions for dealing with partially-labelled datasets during training, there are several limitations that should be considered. The first is the use of only three datasets in our main experiments. Increasing the number of datasets and labelling protocols would enable us to further validate our key findings.
In addition, the annotations in the three datasets have variable quality. It has been shown in previous studies that annotation differences can be picked up by the models during training \citep{leclerc_deep_2019}, which could lead to some conflict when combining these datasets. Indeed, our experimental analysis of annotation differences between CAMUS and Unity Imaging (see Section \ref{sec:expt3_anndiff}) showed a moderate level of annotation discrepancy.

\subsection{Future work} 
To address the challenges identified, future work could include increasing the number of datasets and labelling protocols used for experimental analysis.
The impact of annotation differences on the partial labelling loss functions should also be investigated.
Furthermore, in this paper we only evaluated loss function based solutions to the problem of training with partially-labelled datasets, therefore the work could also be extended to exploring other types of solutions, such as architectural approaches \citep[]{billot2024network, dmitriev2019learning, zhang2021dodnet} which could potentially improve the segmentation results. Additionally, a quality assessment step could be implemented before model training to ensure that only good quality annotations are used. \revision{Also, although dice-based evaluation demonstrates strong geometric agreement and provides a standard measure of segmentation quality, further evaluation in terms of downstream clinical metrics such as LVEF and ventricular volumes would provide a more comprehensive assessment of clinical utility.}

\section{Conclusion}
To summarise, the key conclusions of the comparative investigation presented in this report are:

\begin{itemize}
    \item All three loss functions (aCCE, marginal and aBCE) perform similarly well for an intra-domain partial-labelling segmentation problem.
    \item The aCCE loss experienced a drop in performance in the inter-domain setting, but the performance gap was reduced by the use of label dropout.
    \item Including even a relatively small amount of (partially labelled) data from a new dataset can have a beneficial effect on model performance, and reduce the need for large amounts of training data.
    \item Taking into account the above conclusions, partial labelling loss functions have the potential to train robust and generalisable models for echo segmentation using multiple diverse datasets.
\end{itemize}


\acks{This work was supported by funding from the EPSRC Centre for Doctoral Training in Medical Imaging (EP/L015226/1).}

%
\ethics{The use of the GSTFT database was approved by the London Research Ethics Committee (11/LO/1232), all patients provided written informed consent for participation and the research was conducted according to the Helsinki Declaration guidelines on human research.}

\coi{We declare we don't have conflicts of interest.}

\data{The datasets supporting the findings of the study are publicly available online.}

\bibliography{latex/sample}





\end{document}